\documentclass{bmvc2k}

\title{An Efficient Integration of Disentangled Attended Expression and Identity Features For Facial Expression Transfer and Synthesis }

\addauthor{Kamran Ali}{kamran@knights.ucf.edu}{1}
\addauthor{Charles E. Hughes}{ceh@cs.ucf.edu}{1}

\addinstitution{
 Synthetic Reality Lab\\
 University of Central Florida, Orlando\\
 Florida, USA
}

\runninghead{Student, Prof, Collaborator}{BMVC Author Guidelines}

\begin{document}

\maketitle

\begin{abstract}
In this paper, we present an Attention-based Identity Preserving Generative Adversarial Network (AIP-GAN) to overcome the identity leakage problem from a source image to a generated face image, an issue that is encountered in a cross-subject facial expression transfer and synthesis process.  Our key insight is that the identity preserving network should be able to disentangle and compose shape, appearance, and expression information for efficient facial expression transfer and synthesis.   Specifically,  the expression encoder of our AIP-GAN disentangles the expression information from the input source image by predicting its facial landmarks using our supervised spatial and channel-wise attention module.  Similarly, the disentangled expression-agnostic identity features are extracted from the input target image by inferring its combined intrinsic-shape and appearance image employing our self-supervised spatial and channel-wise attention mod-ule.  To leverage the expression and identity information encoded by the intermediate layers of both of our encoders, we combine these features with the features learned by the intermediate layers of our decoder using a cross-encoder bilinear pooling operation. Experimental results show the promising performance of our AIP-GAN based technique.
 
\end{abstract}
%\vspace{-0.4cm}
%\vspace*{-0.4cm}
%-------------------------------------------------------------------------
\section{Introduction}
%\vspace{-0.2cm}
\label{sec:intro}
Facial Expression synthesis and manipulation by transferring the expression information from a source image to the identity of a target image has recently gained a great deal of attention from the computer vision research community. It is due to the exciting research challenges that it offers apart from its many applications, e.g., facial animation, human-computer interactions, entertainment, and facial reenactment ~\cite{r1}. However, despite promising progress in this domain, synthesizing an expression image by transferring an expression from one image to the identity of another image is still a challenging problem.

Recent facial expression manipulation and editing techniques employ deep generative models to generate synthesized expression images ~\cite{r8}, ~\cite{r9}, ~\cite{r10}, ~\cite{r11},~\cite{r50}. In \cite{r8} an expression controller module is trained using a GAN-based architecture to generate expression images of various intensities. Similarly in \cite{r12}, a unified GAN framework is used to transfer expressions from one image domain to another. Pumarola et al.~\cite{r15} and Shao et al. \cite{r9} exploited Facial Action Units (AU), and the expression synthesis process is guided by the learned AU features. Similarly, in \cite{r13}, \cite{r14}, \cite{r38} and \cite{r79} expression image generation is conditioned on facial landmarks.

Existing facial expression synthesis techniques have the capability to transform the expression of a given image; however, there are three main problems with these methods: 1) During inference they require manual auxiliary information in the form of an expression code, facial landmarks, and/or action unit information to synthesize an expression image. 2) Most of these methods fall short in preserving the identity information of an unseen face image during inference since they fail to disentangle expression features from identity representation. Hence, during the facial expression transfer process, the identity information of the source image is usually leaked through the expression feature vector, which degrades the identity of generated images \cite{r46} 3) Many of these techniques assume a neutral expression based target face image for facial expression transfer. These bottlenecks restrict the practical applicability and efficiency of the facial expression transfer and synthesis process. 

In order to overcome the above-mentioned problems, we present an Attention based Identity Preserving Generative Adversarial Network (AIP-GAN) to automatically and explicitly extract a disentangled expression representation from a source image and disentangled identity features from a target image, having any expression. We then efficiently integrate these features to synthesize a photo-realistic expression image without requiring any auxiliary information such as expression or identity code, facial landmarks or action units during the inference phase, while also preserving the identity of the target image. The learned continuous multi-dimensional expression embedding captures significant variations even within the same semantic category, for instance, modeling various forms of smile expression like a nervous smile, shy smile, laughter, etc. Similarly, the extracted expression-agnostic identity information encodes not only the appearance information of the target face, but it also contains the intrinsic-shape features necessary for proper reconstruction of an identity preserving expression image. 

AIP-GAN consists of an encoder-decoder based architecture containing two encoders, an expression encoder $G_{es}$ and an identity encoder $G_{et}$, each equipped with dedicated spatial and channel-wise attention modules to infer the intrinsic components (such as facial landmarks, shape and textures maps) of a facial expression image. To alleviate the identity leakage problem, we use attention modules in encoder $G_{es}$ to extract only expression related features from the source image by predicting its facial landmarks. Similarly, we use a self-supervised spatial and channel-wise attention mechanism in our identity encoder $G_{et}$ to filter out the expression information of the target image and focus on its appearance and intrinsic-shape information by inferring the combined appearance and shape image of that particular identity. We term that combined appearance and intrinsic-shape image an expression-free identity map and denote it by $H$. To transfer the expression information from the source image to the identity of the target face while synthesizing a realistic-looking identity preserving  expression image, we have designed a novel decoder network that effectively integrates the disentangled attended expression and identity features from intermediate layers of both encoders with the features from intermediate layers of our decoder using the compact bi-linear operation specified in \cite{r78}. 

The main contributions of this paper can be summarized as follows: 1) We develop a novel Attention based Identity Preserving Generative Adversarial Network (AIP-GAN) to efficiently synthesize an expression image by transferring expression information from a source image to a target face while preserving the target's identity information. 2) We employ dedicated spatial and channel-wise attention modules to disentangle intrinsic facial components such as facial landmarks, shape and appearance maps of an expression image, and demonstrate its effectiveness in solving the identity leakage problem from the source image to the target image during the synthesis process. 3) To leverage the attended disentangled expression and identity information encoded by the intermediate layers of our encoders, our novel decoder architecture effectively integrates this encoded information with the features learned by its intermediate layers. 4) Extensive experiments performed under extreme conditions where the identities of the source image and the target image are completely different in terms of facial shape, appearance, and expression, show that our method can be used to transfer not only trivial expressions like smiling and anger, but it has the capability to transfer all six basic expressions while preserving the identity of the target image. Most importantly, our AIP-GAN method outperforms state-of-the-art facial expression transfer and synthesis techniques in terms of preserving the identity of the generated images while transferring expressions.
%\vspace{0.45cm}
\section{Related Work}
%\vspace{-0.3cm}
%\textbf{Facial Expression Manipulation}:
\subsection{Facial Expression Manipulation}
Recently proposed facial expression synthesis methods are mostly based on conditional Generative Adversarial Networks (cGANs) \cite{r35}, \cite{r8},~\cite{r15},~\cite{r36}. Some earlier techniques such as \cite{r35} and \cite{r36}, used deterministic target expressions as one-hot vectors and generated synthesized images conditioned on discrete facial expressions. Ding et al. \cite{r8} proposed an Expressive GAN (ExprGAN) to synthesize an expression image conditioned on a real-valued vector that contains more complex information such as intensity variation. Similarly in \cite{r13} and \cite{r38}, the image synthesis process is conditioned on geometry information in the form of facial landmarks. Choi et al. \cite{r39} proposed the StarGAN method to employ domain information and generate an image into a corresponding domain. In another work, Pumarola et al. \cite{r15} used AUs as a conditional label to synthesize an expression image. All of the above techniques rely on explicit information such as expression, AU and facial landmarks to synthesize an expression image during the inference stage.
%\textbf{Attention based Disentanglement}:
\vspace*{-0.4cm}
\subsection{Attention based Disentanglement}
Many previous facial manipulation techniques \cite{r42}, \cite{r50}, \cite{r59}, \cite{r60}, \cite{r61}, \cite{r62}, \cite{r63} have used disentanglement methods for feature extraction to transfer facial attributes such as expressions, eyeglasses, facial hair, etc. to a target image. Similar to our disentanglement by decomposition technique, \cite{r42}, \cite{r59}, \cite{r42}, \cite{r60}, \cite{r62} decompose an input image to components such as shape and appearance to edit a facial expression image. However, these techniques extract features by directly operating on the entire image, and thus inevitably obtain highly entangled features from irrelevant regions of the image. To overcome this problem, Zhang et al. \cite{r64} employed a spatial attention mechanism for the extraction of features from attribute-relevant regions of an image to effectively synthesized images. However, since their architecture is not explicitly designed for facial expression transfer and synthesis, they have validated their technique only on simple expressions such as smile and surprise. In this paper, we demonstrate that a proper disentanglement of relevant features can be achieved by using dedicated spatial and channel-wise attention modules, which proves to be effective for identity preserving facial expression transfer and synthesis.  

% Figure 1 *************************
\begin{figure}[ht!]
%\vspace{-3cm}
\centering
%[scale=1, width=.01\textwidth]
\includegraphics[width=13cm,, height=12.cm]{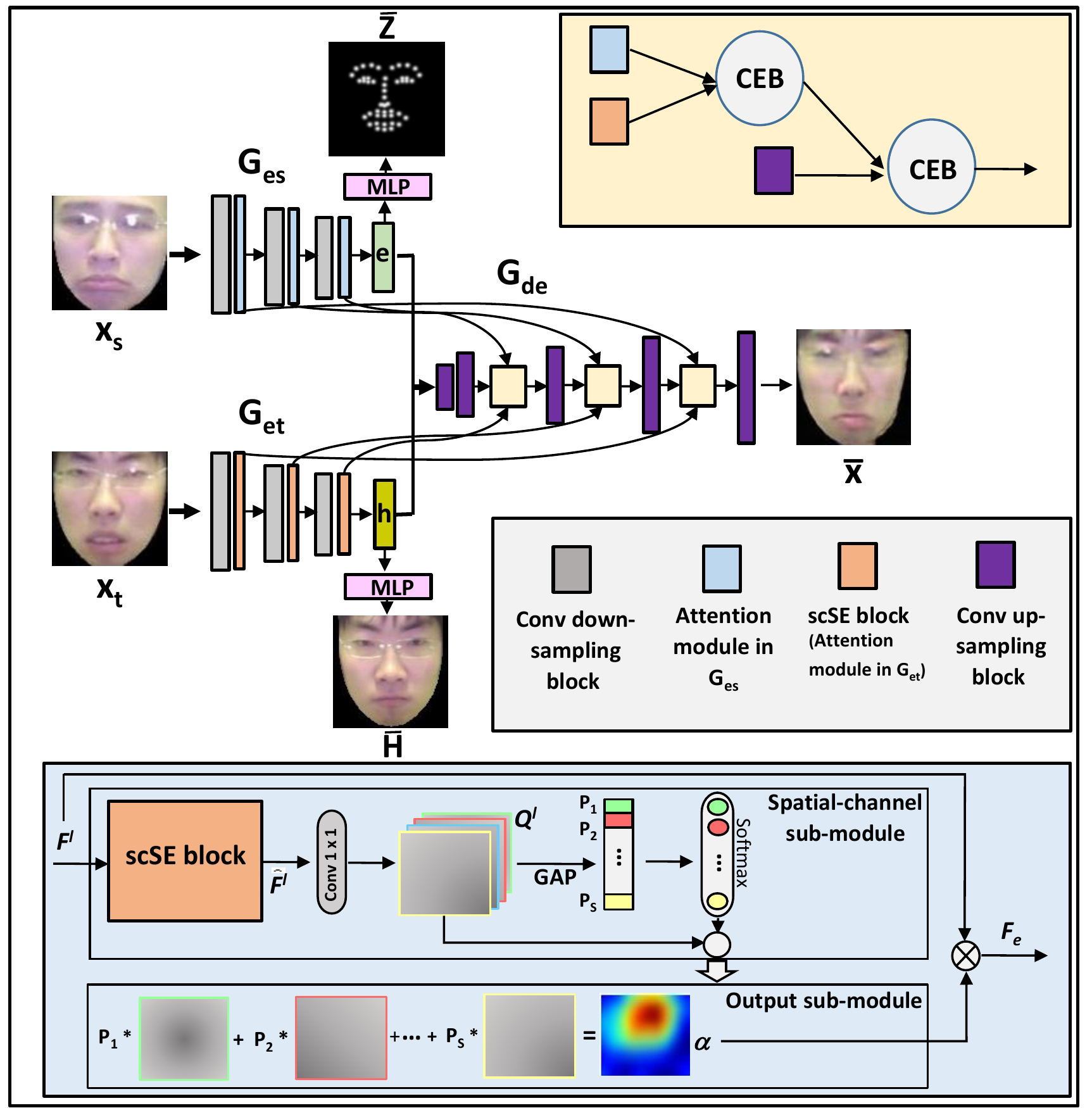}
\caption{\textbf{An overview of our AIP-GAN (best viewed in color):} Given a source image, $x_s$ and a target image, $x_t$ as input, $G_{es}$ disentangles and extracts expression information from $x_s$ by reconstructing the facial landmark image $\bar{Z}$ using our supervised spatial-channel attention modules (shown in light blue color), and $G_{et}$ extracts expression-agnostic identity information from $x_t$ by synthesizing its combined shape and appearance map $\bar{H}$ employing Spatial and Channel Squeeze $\&$ Excitation Blocks (scSE) \cite{r57} (shown in brown color). Our decoder $G_{de}$ combines the features learned by its intermediate layers with the disentangled attended expression and identity features from the intermediate layers of both the encoders by using cross-encoder bilinear  operation (CEB) modules (shown in yellow color).}
\label{fig:1}
%\vspace{-0.66cm}
\end{figure}
%\vspace{-0.65cm}
\section{Overview of AIP-GAN}
\vspace{-0.35cm}
Given a source image $x_s \in{X}$ and a target image $x_t\in X$, the main objective of AIP-GAN is to transfer the expression of $x_s$ to the identity of $x_t$. Specifically, an encoder $G_{es}: X\rightarrow{E}$ is used to encode the expression representation $e \in E $ from $x_s$, and an encoder $G_{et}: X\rightarrow{H}$ is employed to encode the expression-agnostic identity representation $h \in H $ from $x_t$. A decoder $G_{de}: E\times H \rightarrow{X}$ is then used to map the expression and the identity latent embedding space back to the face space. The synthesized expression image $\bar{x}$ is generated by computing $\bar{x} = G_{de} (G_{es}(x_s),G_{et}(x_t))$.
\vspace{-0.45cm}
\subsection{Attention based Disentanglement and Synthesis Framework}
%\vspace{-0.25cm}
Figure \ref{fig:1} shows the overall framework of our AIP-GAN. The network $G_{es}$ disentangles, and extracts the expression representation $e$ from $x_s$ by using dedicated supervised attention modules to predict its facial landmark image, and the model $G_{et}$ encodes the identity information to a representation $h$ from $x_t$ by employing self-supervised spatial and channel-wise attention modules to reconstruct the expression-free identity map $\bar{H}$. A realistic expressive face is then constructed progressively by combining the source expression and the target identity features. Specifically, the decoder $G_{de}$ is designed by having skip connections like U-Net \cite{r65} from both $G_{es}$ and $G_{et}$ leveraging the disentangled attended features to  generate a realistic-looking expression image while preserving the encoded information.\\
\textbf{Expression Encoder ($G_{es}$)}: The expression encoder $G_{es}$ is a multi-task network with two objectives: to predict the expression label of the source image $x_s$ and to reconstruct its facial landmark image $\bar{Z}$. To achieve these objectives, we employ dedicated spatial and channel-wise attention modules in our expression encoder. Inspired by the supervised attention mechanism proposed in \cite{r65}, we train our expression attention modules in a supervised manner. However, we enhanced the effectiveness of the attention module in \cite{r65} by combining it with the Concurrent Spatial and Channel Squeeze $\&$ Excitation Block (scSE) of \cite{r57}. We divide each expression attention module into two sub-modules: spatial-channel sub-module and output sub-module, as shown in Figure \ref{fig:1}. Inside the spatial-channel sub-module, the scSE block takes the $l^{th}$ layer feature activation $F^l \in R^{C\times H \times W}$ as input and outputs attended feature maps $\hat{F^l} \in R^{C\times H \times W}$, where C, H and W are the number of channels, the height, and the width of feature activations, respectively. $\hat{F^l}$ is then passed through $1 \times 1$ conv layer to produce feature maps $Q^l\in R^{S\times H \times W}$ with a reduced dimension, i.e., $S < C$, where $S=6$ corresponding to six basic expressions. $Q^l$ is passed through a global average pooling layer and a softmax layer successively to produce a probability vector $P$, which is used in the output sub-module to produce a final attention map $\alpha$ by performing a weighted sum operation on the 2-D feature activations of $Q^l$ according to their confidence scores determined by the entries of $P$, i.e., $\alpha = norm(\sum_{i=1}^S P_i Q_i)$, where $norm$ denotes the normalization operation. Finally, $\alpha$ is element-wise multiplied with ${F^l}$ by broadcasting to produce attended expression features $F_e = {F^l}\odot\alpha$, which act as input to the next encoder layer. The attention loss of the $l^{th}$ layer can be represented by the following unified formula:
%\vspace{-0.2cm}
\begin{align}
\mathcal{L}_{att}^{l}=-\frac{1}{N}{}&\sum_{n=1}^N \sum_{i=1}^S 1[g_n=i] \log{P_i}
\end{align}
%\vspace{-0.2cm}
In this formula, $1[f]= 1$ if condition $f$ is true, and 0 otherwise. $N$ corresponds to the number of input images, and $g_n$ denotes the expression label of the $n^{th}$ input image. We employ attention modules in all convolutional layers of our expression encoder and train them with equal weights assigned to their attention loss.

We use a simple MLP to reconstruct $\bar{Z}$ from $e$ by employing a linear transformation, as $x_s$ and ${Z}$ both are registered images. The expression encoder focuses on important regions of a face such as eyes, mouth and nose to extract expression relevant features by predicting facial landmarks, which is governed by the formula: $\mathcal{L}_{d}=||\bar{Z}(x_s)-Z(x_s)||_2$, where $\bar{Z}(x_s)$ and $Z(x_s)$ are the predicted and ground truth facial landmark images, respectively.\\
\textbf{Identity Encoder ($G_{et}$)}: To generate an expression image that preserves the identity of the target image $x_t$, our encoder $G_{et}$, as opposed to previous techniques, where identity features are extracted by focusing only on the appearance information, learns both the intrinsic-shape and appearance features of the target image. In particular, $G_{et}$ maps the input target image $x_t$ to an expression-agnostic identity representation $h$ by using the self-supervised scSE block of \cite{r57} as an attention module in its intermediate layers. The expression-free identity image $\bar{H}$ is then reconstructed from $h$ by using a fully connected layer based linear transformation, as both $x_t$ and $H$ are registered images. We generate the ground truth identity map $H$ by modifying the AMM \cite{r74}, \cite{r75} algorithm in such a way that the mean shape $s_o$ is replaced by the mean of neutral images of each individual in the dataset. The appearance information defined within the base mesh $s_o$ is then generated by modeling only the relevant pixels. Therefore, $G_{et}$ extracts expression-agnostic identity features from $x_t$ by reconstructing $\bar{H}$ from $h$, which is guided by the given loss function: $\mathcal{L}_{h}=||\bar{H}(x_t)-H(x_t)||_2$, where $\bar{H}(x_t)$ and $H(x_t)$ are the predicted and ground truth identity maps, respectively.\\
\textbf{Decoder ($G_{de}$)}: To leverage the semantic information encoded in the intermediate layers of our encoders, our decoder architecture is similar to U-Net \cite{r65}, as shown in Figure \ref{fig:1}. The attended expression and identity feature maps are skip-connected with the corresponding intermediate layers of $G_{de}$ to transfer the expression of the source image and preserve the identity of the target face in the generated image. To effectively integrate the expression and identity features, and features of the intermediate layers of $G_{de}$, we use the compact bilinear pooling operation proposed in \cite{r78}, referring to this as a cross-encoder bilinear (CEB) pooling module. The CEB module models the interactions of features by utilizing the pairwise correlations developed between the channels. 
%\vspace{-0.35cm}
\subsection{Network Training}
For the training of our network, we adopt the curriculum learning strategy of \cite{r8}, \cite{r46}, where we first train the expression encoder and the identity encoder separately to learn the expression and identity embedding by reconstructing the facial landmark images and the combined shape and appearance images, respectively, using $\mathcal{L}_{d}$, $\mathcal{L}_{att}$, and $\mathcal{L}_{h}$. We then attach both the encoders with the decoder of our generator to reconstruct expression images using the expression and identity embedding. During this stage of training, we freeze both encoders and only train the decoder using $\mathcal{L}_{if}$, $\mathcal{L}_{irec}$, and $\mathcal{L}_{ef}$. In the final stage of our training, we connect the discriminator with the generator and jointly train the whole AIP-GAN network to refine the image to be photo-realistic by optimizing the total loss function given below:   
%\vspace{-0.25cm}
\begin{align}
\mathcal{L}_{total}={}&\lambda_1\mathcal{L}_{irec}+\lambda_2\mathcal{L}_{if}+\lambda_3\mathcal{L}_{ef} + \lambda_4\mathcal{L}_{d} + \lambda_5\mathcal{L}_{h}+\lambda_6\mathcal{L}_{att}+\lambda_7\mathcal{L}_{adv}
\label{eq:2}
\end{align}
Where $\mathcal{L}_{irec}$ and $\mathcal{L}_{if}$ corresponds to ${L}_{pixel}$ and ${L}_{id}$ loss of \cite{r8}, respectively, computed between $x_t$ and $\bar{x}$. $\mathcal{L}_{ef}$ is the ${L}_{id}$ loss of \cite{r8} between $x_s$ and $\bar{x}$ to preserve the semantic (expression) information in the synthesized image $\bar{x}$. To compute this loss we train the VGG face model \cite{r66} on AffectNet dataset \cite{r67}. $\mathcal{L}_{adv}$ is the multi-scale GAN loss \cite{r68}. During the joint training of the network, we empirically set $\lambda_1= 20,~\lambda_2,~\lambda_3 = 10,~\lambda_4,~\lambda_5=5,~\lambda_6,~\lambda_7 = 3$. 
%\vspace{-0.35cm}
\section{Experiments}
%\vspace{-0.25cm}
%\textbf{Datasets}: 
\subsection{Datasets and Implementation Details}
We evaluate AIP-GAN by conducting experiments on three popular facial datasets: Oulu-CASIA \cite{r71}, BU-4DFE \cite{r72} and CelebA \cite{r73}. Oulu-CASIA (OC) dataset contains 480 video sequences of 80 subjects, exhibiting six basic expressions. In this experiment, only images captured under a strong condition with VIS camera are used. From each video sequence, we extract the starting neutral frame and the last three frames corresponding to peak expression to construct the dataset. The BU-4DFE dataset consists of 60,600 images collected from the video sequences of six basic expressions of 101 identities. During our experiments, we extract the starting neutral image and the three middle images corresponding to the peak expression in a video sequence. CelebA contains 202,599 face images of celebrities, and it is a widely used large-scale facial dataset. All the experiments are performed by splitting the dataset into training and testing sets in a person independent way, with proportions of $90\%$ and $10\%$, respectively. To assess the generalization capability of our proposed method, we evaluated our technique on two different testing data: 1) data from the same distribution, and 2) data from an \textquotedblleft in the wild\textquotedblright cross dataset validation in which the network is trained on the BU-4DFE dataset and validated with the frontal images selected from the CelebA dataset.\\
\textbf{Pre-processing:} Facial landmarks of input face images are obtained by using the Dlib Library \cite{r69}. Based on these facial landmarks we generate the ground truth combined shape and appearance identity image $H$ by employing the combined AMM algorithm \cite{r74}, \cite{r75}. All face images are aligned, cropped and resized to $96 \times 96$. To alleviate overfitting, we also performed random flipping of the input images to augment the dataset.\\
\textbf{Implementation Details}: AIP-GAN is implemented using PyTorch, and trained using the Adam optimizer \cite{r70}, with a batch size of 64 and initial learning rate of 0.0002.

%\textbf{Network Architecture}: 
\subsection{Network Architecture}
The architecture of both encoders, $G_{es}$ and $G_{et}$, is designed based on seven downsampling blocks, with each block consisting of a $3\times3$ stride 1 convolution, instanceNorm and LeakyReLU, and each downsampling block is followed by an attention module. The number of channels in both $G_{es}$ and $G_{et}$ is 64, 128, 128, 256, 256, 512, 512, and one FC layer for the expression feature vector $e$ (30-dimensional), and one FC layer for the identity feature vector $h$ (50-dimensional), respectively. To reconstruct the facial landmark image $\bar{Z}$ (from $e$) and the expression-free identity image $\bar{H}$ (from $h$), we use two shallow multi-layer perceptrons with ReLU non-linearity. The decoder $G_{de}$ is built on seven upsampling blocks having the number of channels: 512, 512, 256, 256, 128, 128, 64, and the first, third, fifth and sixth block containing [cross-encoder bilinear pooling (CEB) module, convolution, upsample], and the rest of the blocks contain convolution and upsample layers.
%\vspace{-0.65cm}

% Figure 2 ***************************
\begin{figure}
\centering
%[scale=1, width=.01\textwidth]
\includegraphics[width=13cm,, height=8cm]{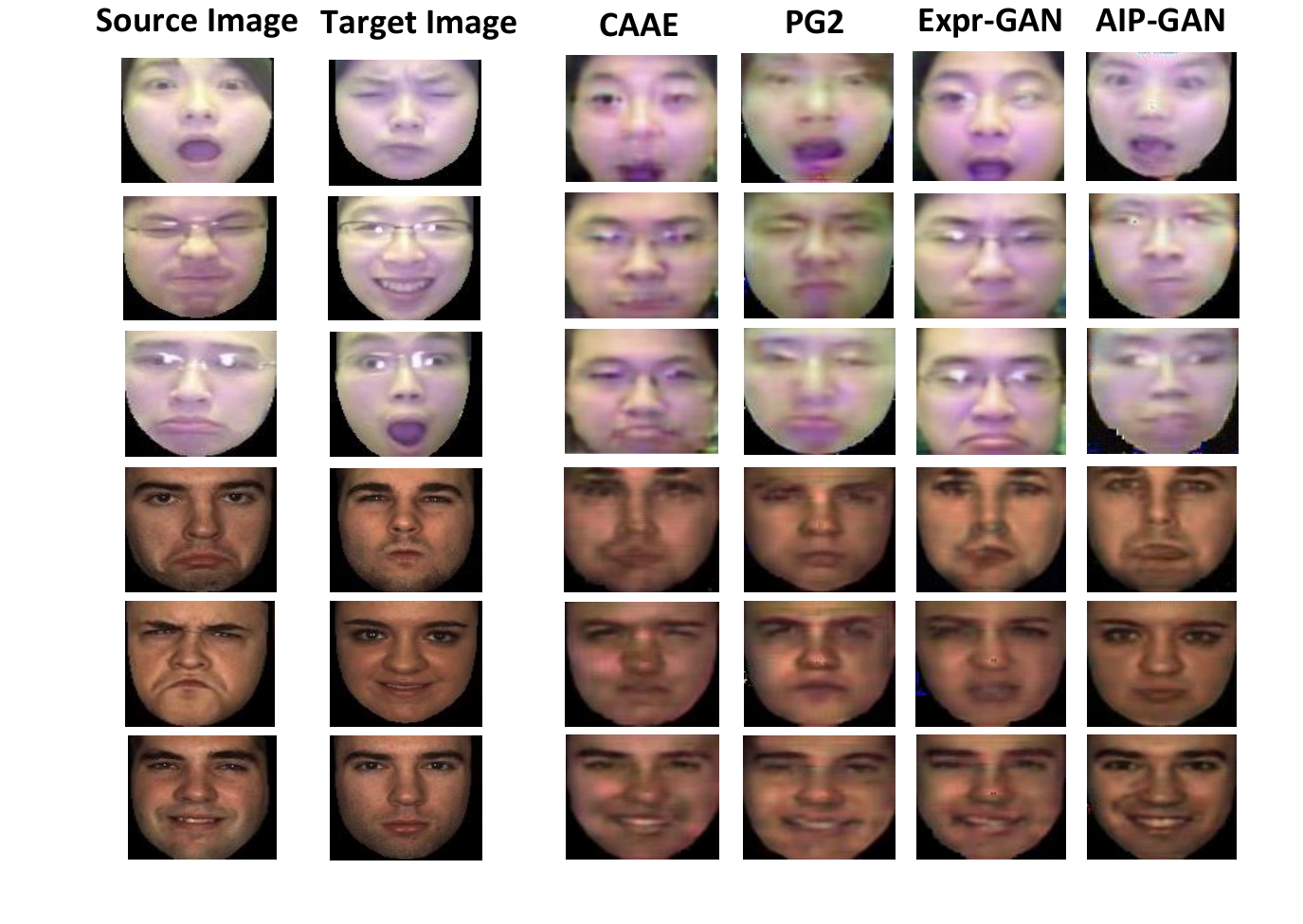}
\vspace{-1.0cm}
\caption{\textbf{Qualitative comparison with-state-of-the-art image based facial expression synthesis techniques:} Our AIP-GAN generates expression images by transferring the expression of source images to the face of target images while preserving the identity of the target image in a better and more consistent way as compared to CAAE \cite{r77}, Expr-GAN \cite{r8} and PG2\cite{r81}. Note that the input target images contain non-neutral expressions as well.}
%\vspace{-0.45cm}
\label{fig:2}
\end{figure}

% Figure 3 *****************************
%\vspace{-0.70cm}
\subsection{Comparison with State-of-the-art Techniques}
\begin{figure}
\centering
%[scale=1, width=.01\textwidth]
\includegraphics[width=12cm,, height=7cm]{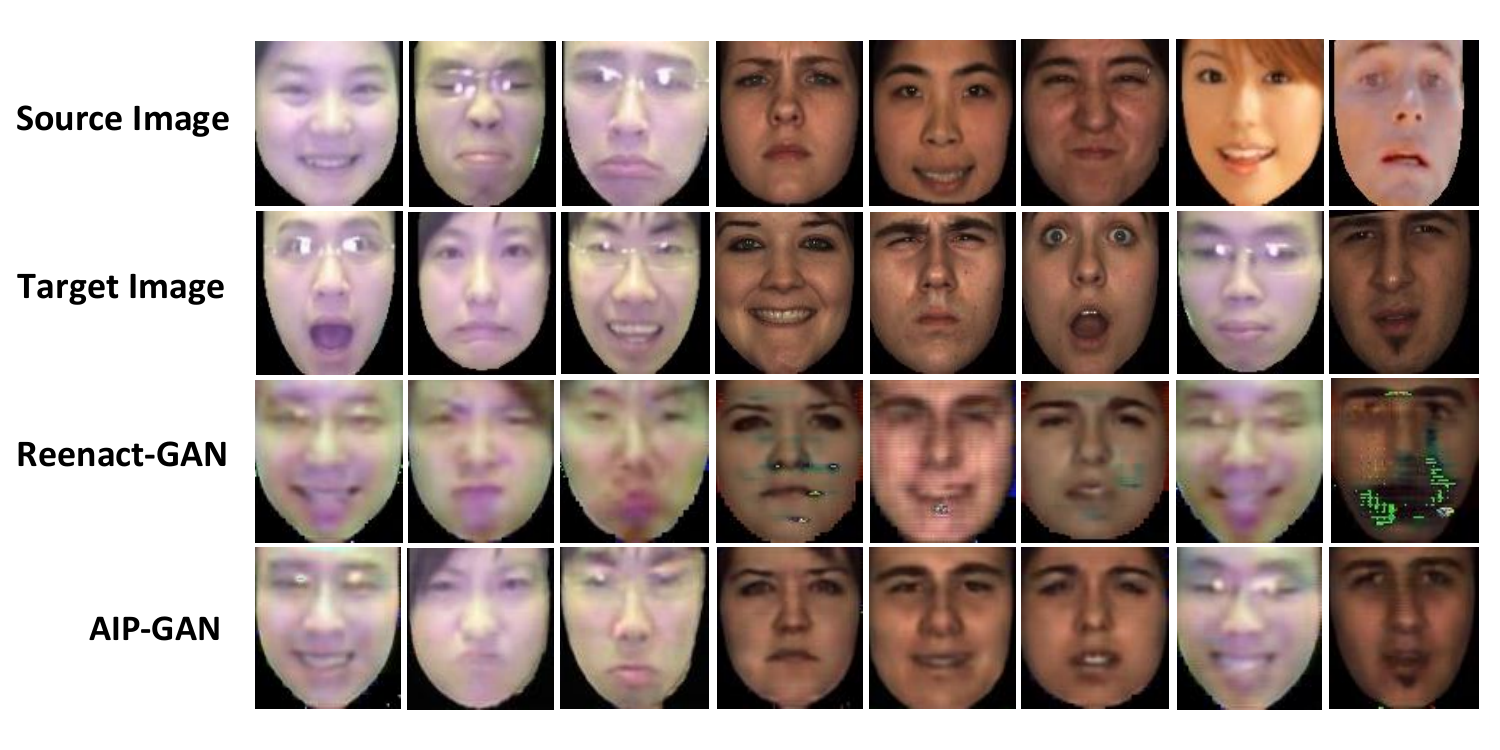}
\vspace{-0.6cm}
\caption{\textbf{Qualitative comparison with-state-of-the-art video based ReenactGAN \cite{r82}:} ReenactGAN fails to perform facial expression transfer and synthesis effectively when trained on limited frames available per identity. AIP-GAN, on the other hand, generates a realistic-looking expression image by transferring the expression of a source image to the face of a target image while preserving the identity of the target image.}
\label{fig:3}
%\vspace{-0.6cm}
\end{figure}

%\vspace{-0.2cm}
In order to evaluate the performance of our method we compare our results with two types of facial expression manipulation techniques, i.e, image based methods and video based state-of-the-art face reenactment algorithm ReenactGAN \cite{r82}.\\
\textbf{Image based methods}: For our comparison with image based methods, we select two state-of-the-art facial manipulation techniques: ExprGAN \cite{r8} and CAAE \cite{r77}, and another network from the domain of person image generation, PG2 \cite{r81}. Figure \ref{fig:2} shows the result of image based techniques and our proposed AIP-GAN method. Expr-GAN generates good quality expression images, but it suffers from the identity shift problem and thus cannot preserve the identity of the target image. However, Expr-GAN transfers the expression of the source image to the target identity quite efficiently, but the downside is that this expression transfer process is trained not in an end-to-end manner. PG2 also fails to preserve the identity of the target image while transferring the expression of the source image. Similarly, the expression images generated by CAAE contain artifacts, and they also face the identity shift problem. The expression images generated by our AIP-GAN method transfers the expression of the source image to the target image while preserving the identity of the target face. Note that the expressions of the input target images are non-neutral expressions, which provides strong evidence for the efficacy of the expression-agnostic identity feature extraction process of our identity encoder.\\
\textbf{Video based method}: ReenactGAN \cite{r82} is a video based face reenactment method that is trained on a video of a target identity to properly model the identity information of the target face. However, in practical applications, it is very difficult to collect large number of high quality frames containing the face of the target identity to reenact a new person. In contrast, our proposed method effectively modifies the expression of an unseen person while preserving its identity with just a single input image. Figure \ref{fig:3} shows that ReenactGAN produces blurred and distorted expression images when trained on the limited number of frames present in video sequences of identities present in the datasets. On the other hand, our image-based AIP-GAN generates good quality expression images by transferring the expression of the source image to the target face, while preserving its identity. 

%\vspace{-0.55cm}
\subsection{Quantitative Results}
%\vspace{-0.25cm}
We quantitatively compare our method with the aforementioned techniques from two perspectives: 1) the quality of generated expression images in terms of preserving the identity of target images, and 2) the capability of transferring expression information from source to target images. We validate the identity preserving capability by employing VGG face model \cite{r66} to conduct face verification on the generated faces. The threshold for the cosine similarity is set to 0.68. Similarly, in order to evaluate the effectiveness of our method in transferring the expression of source images to target images, we use VGG face model \cite{r66} trained on AffectNet dataset \cite{r67}, and measure the cosine similarity between the output of its last convolution layer. The quantitative comparison result given in Table \ref{table:1} shows the significance of using our attention mechanism combined with CEB modules in the decoder layers. From the face verification columns, we can observe that our framework outperforms the state-of-the-art image and facial expression manipulation techniques in terms of preserving the identity information of the generated images. High expression similarity scores are obtained in the case of PG2, and we believe that it is due to the reason that the network is trained with paired data, which leads to better expression supervision. Our expression similarity values are quite close to the expression similarity scores obtained from Expr-GAN, which employs an off-the-shelf expression recognition network to get the expression label that is then fed to Expr-GAN as a conditional vector. On the other hand, our AIP-GAN performs the expression transfer and synthesis process in an end-to-end manner by extracting the expression information automatically from the source image. 

\begin{table}
\begin{center}
\begin{tabular}{| *{5}{c|} }
    \hline
    & \multicolumn{2}{c|}{Same distribution}
            & \multicolumn{2}{c|}{Cross-data validation(in the wild)}\\
    \hline
Method   &   Face-Verf  &  Exp-Sim &   Face-Verf  &   Exp-Sim   \\
    \hline\hline
CAAE \cite{r77} & 0.616 & 0.795 & 0.548 & 0.737 \\
\hline
PG2 \cite{r81}&   0.709& \textbf{0.927} & 0.594 & \textbf{0.906} \\
    \hline
Expr-GAN\cite{r8}&   0.849& \textbf{0.919} & 0.761 & 0.828 \\
    \hline
ReenactGAN \cite{r82}&   0.832& 0.867 & 0.752 & 0.773 \\
\hline
Ours w/o $\mathcal{L}_{att}$ and CEB &   0.816& 0.841& 0.725 & 0.753\\
\textbf{Ours}   & \textbf{0.907}& 0.903& \textbf{0.856} & 0.841\\
\hline
\end{tabular}
\end{center}
%\vspace{-0.2cm}
\caption{\textbf{Quantitative comparison:} We evaluate the identity preserving ability of each method using the face verification technique. Here, the average expression similarity score of each method indicates its ability to transfer expression information from source images to target faces. These two metrics are evaluated on test data from two groups: the same data distribution and an \textquotedblleft in the wild\textquotedblright cross data distribution.}
\label{table:1}
%\vspace{-0.3cm}
\end{table}

%\vspace{-0.7cm}
\section{Conclusion}
%\vspace{-0.45cm}
In this paper, we present an Attention Identity Preserving Generative Adversarial Network (AIP-GAN) for synthesising an expression image by transferring expressions across different identities. Different from conventional expression synthesis techniques in which expression synthesis is carried out by using pre-defined emotion labels, AIP-GAN extracts facial expressions in a continuous fashion by learning a compact expression embedding. Specifically, expression information from the source image is transferred to the target image by preserving the identity of the target image using an attention based encoder-decoder GAN architecture. AIP-GAN disentangles the expression and identity information from the input images by inferring the representations of intrinsic components of face expression images, including facial landmarks, shape, and texture maps. Experimental results demonstrate that our technique can synthesize identity preserving realistic looking expression images by transferring the expression of the source image to the target face even when there exist significant differences in facial shapes and expressions between the two input images.

\bibliography{egbib}
\end{document}